\documentclass[10pt,twocolumn,letterpaper]{article}

\usepackage{iccv}
\usepackage{times}
\usepackage{epsfig}
\usepackage{graphicx}
\usepackage{amsmath}
\usepackage{amssymb}
\usepackage{booktabs}
\usepackage{multirow}
\usepackage{microtype}

\usepackage{float}
\usepackage{xcolor}
\usepackage{subcaption}
\usepackage[sort&compress,numbers,square,comma]{natbib}

\usepackage[pagebackref=true,breaklinks=true,letterpaper=true,colorlinks,bookmarks=false]{hyperref}

\iccvfinalcopy %

\def\iccvPaperID{6308} %

\ificcvfinal\fi

\def\mybox#1{\leavevmode \setbox0=\hbox{\framebox{#1}}%
   \dimen0=\wd0 \edef\posxA{\expandafter\ignorept\the\dimen0 \space}%
   \hbox{\kern3pt\pdfliteral{q .8 .8 1 rg .8 .8 1 RG .9963 0 0 .9963 0 0 cm 1 j 1 J 6 w
                             0 0 m 0 5 l \posxA 5 l \posxA 0 l 0 0 l B Q}%
         \box0 \kern3pt}%
}
{\lccode`\?=`\p \lccode`\!=`\t  \lowercase{\gdef\ignorept#1?!{#1}}}

\begin{document}

\title{Self-Denoising Neural Networks for Few Shot Learning}

\author{Steven Schwarcz\\
University of Maryland, College Park\\
College Park, MD\\
{\tt\small sasz11@gmail.com}
\and
Saketh Rambhatla\\
University of Maryland, College Park\\
College Park, MD\\
{\tt\small rssaketh@umd.edu}
\and
Rama Chellappa\\
Johns Hopkins University\\
Baltimore, MD\\
{\tt\small rchella4@jhu.edu}
}

\maketitle
\ificcvfinal\thispagestyle{empty}\fi

\begin{abstract}
In this paper, we introduce a new architecture for few shot learning, the task of teaching a neural network from as few as one or five labeled examples. Inspired by the theoretical results of Alaine \etal that Denoising Autoencoders refine features to lie closer to the true data manifold, we present a new training scheme that adds noise at multiple stages of an existing neural architecture while simultaneously learning to be robust to this added noise.  This architecture, which we call a Self-Denoising Neural Network (SDNN), can be applied easily to most modern convolutional neural architectures, and can be used as a supplement to many existing few-shot learning techniques. We empirically show that SDNNs out-perform previous state-of-the-art methods for few shot image recognition using the Wide-ResNet architecture on the \textit{mini}ImageNet, \textit{tiered}ImageNet, and CIFAR-FS few shot learning datasets. We also perform a series of ablation experiments to empirically justify the construction of the SDNN architecture. Finally, we show that SDNNs even improve few shot performance on the task of human action detection in video using experiments on the ActEV SDL Surprise Activities challenge. 
\end{abstract}

\section{Introduction}

Despite the ubiquitous power of modern deep learning techniques for image and video classification, learning from very few examples still remains a difficult problem. A visual recognition system deployed in a real world setting may need to identify new classes of which it has only seen one or two examples. Learning in this low-data regime, where one might have access to as little as a single labeled example, generally requires entirely different techniques than traditional supervised learning, and the pursuit of these techniques has launched an entire sub-field of machine learning research known as  ``few shot learning''.

Few shot learning is generally characterized by a two-stage approach. In the first stage, the network is pretrained in a supervised setting with a large labelled dataset of known classes. In the second stage, the network is exposed to a limited number of labeled examples in novel classes that were not included in the pretraining dataset, usually with as few as five or even one sample per class. The final system is then evaluated against a held-out test set of images belonging to the novel classes.

Under this paradigm, the machine learning community has developed a large number of techniques. These techniques can themselves be classified into a broad set of categories, ranging from meta-learning techniques that quickly learn to perform meaningful parameter updates when presented with new data \cite{Finn2017ModelAgnosticMF,Rusu2019MetaLearningWL,Vuorio2019MultimodalMM,Nichol2018OnFM,Lee2019MetaLearningWD}, to feature hallucination techniques that generate additional samples for training \cite{Hariharan2017LowShotVR,Wang2018LowShotLF} and transductive techniques that make use of the correlations between novel class samples during inference \cite{Dhillon2020ABF}. In this work, we follow the lead of other metric-based approaches \cite{miniimagenet,NIPS2017_cb8da676,Sung2018LearningTC} which perform classification using highly representative features. That said, our approach is orthogonal enough to many existing metric-based approaches, and simple enough to implement, that it can be easily combined with existing approaches to achieve easy few shot performance gains.

The conceptual backbone of our approach is similar to that of Denoising Auto-Encoders \cite{Vincent2008ExtractingAC} (DAEs), which were shown to be useful for few shot learning by Gidaris \etal \cite{Gidaris_DAE}. DAEs are effective for metric learning because they build robust feature extractors that, by removing noise from features, move features towards their most likely configuration in feature space. At inference time, the noise is not applied and the feature is refined to be more representative, in effect creating a more robust prototype for few shot learning. 

Our approach expands on this concept by making a few key observations. The first is that in order to counteract the effects of noisy features, it is not actually necessary to reconstruct the features precisely. Instead, it is sufficient merely to ensure that the network can accomplish the same classification task it would have preformed otherwise. This can be enforced simply by adding an additional classification loss immediately after the noise is applied. Our second observation is that this denoising process need not be a separate module from the rest of the network applied iteratively, but can instead be integrated directly into existing network training with almost no architectural changes or differences in learning hyperparameters. 

The final result is a relatively straightforward addition to existing neural network architectures. The features of an existing architecture, such as a ResNet \cite{resnet}, are modified before each block with some form of noise. The network is also augmented with auxiliary losses after each block, similar to those used to train \eg Inception \cite{inception}. The end result, which we refer to as a Self-Denoising Neural Network (SDNN), is a network that continually refines its features as the network deepens, ultimately producing features that are altogether more robust than the original architecture by itself.

We also show, through detailed ablation experiments in Section \ref{sec:exp} of this paper, the significance of many of the observations discussed above. For instance, we observe that the auxiliary losses used by our method are essential; merely adding noise to the network as it trains has nearly no effect on performance. It is only by constantly enforcing that the features produce the same evaluation results that denoising produces more robust features.

In summary, our contributions are as follows:

\begin{itemize}
\item We present a novel approach to metric learning for few shot visual tasks termed Self-Denoising Neural Networks (SDNNs). In principle nearly any existing deep architecture for visual recognition can be converted into an SDNN using a few simple modifications to the training pipeline.
\item We demonstrate the effectiveness of SDNNs on the \textit{mini}ImageNet, \textit{tiered}-ImageNet, CIFAR-FS, and ActEV Surprise Activities datasets. We also show that our method is very general, and can be easily added to existing metric-based few shot techniques.
\item We empirically analyze our SDNN architecture through a series of detailed ablation experiments.
\end{itemize}

\section{Related Works}

Few shot learning \cite{Lake2015HumanlevelCL} for deep learning is a rich area of research, with prior work falling into several broad categories. One of the more prominent categories is optimization-based meta-learning \cite{Finn2017ModelAgnosticMF,Rusu2019MetaLearningWL,Vuorio2019MultimodalMM,Nichol2018OnFM,Lee2019MetaLearningWD}. These methods ``learn-to-learn’’ by pretraining models that are able to produce gradient updates that facilitate quick fine-tune on small numbers of samples. Methods like MAML \cite{Finn2017ModelAgnosticMF}, LEO \cite{Rusu2019MetaLearningWL}, and Reptile \cite{Nichol2018OnFM} do this by incorporating the fine-tuning step into the learning stage.

Other metric learning techniques deal with minimal data by hallucinating additional samples for training \cite{Hariharan2017LowShotVR,Wang2018LowShotLF}. Still others improve performance by incorporating more general machine learning techniques, like self-supervised learning \cite{Gidaris_2019_ICCV,Su2020WhenDS} or knowledge distillation \cite{Tian2020RethinkingFI}.

Metric-based few shot learning algorithms are another large family of techniques \cite{Koch2015SiameseNN,miniimagenet,NIPS2017_cb8da676,Oreshkin2018TADAMTD,Sung2018LearningTC,Lifchitz2019DenseCA,Satorras2018FewShotLW,Tseng2020CrossDomainFC}. These methods operate by embedding inputs into a feature space where images of the same class are naturally close together by some metric. For example, RelationNet \cite{Sung2018LearningTC} accomplishes this using cosine similarity, and MatchingNet \cite{miniimagenet} uses Euclidean distance. Though already quite simple, many of these methods \cite{miniimagenet,NIPS2017_cb8da676,Sung2018LearningTC} utilize a pretraining scheme that mimics the few shot training stage by sampling a small support set of sample classes during training. Other metric-based methods like Cosine Classifiers \cite{cosine1,cosine2,chen19closerfewshot} remove even this training complication, providing a methodology for building robust features without the need for support sets during training. In this paper, we design our method on top of these Cosine Classifier based methods, motivated in large part by a pursuit of simplicity.

\section{Method}

In the standard few shot learning problem setup for visual tasks, we are initially given access to a set of images \footnote{Though we use the term image here, $x_*$ could also be a video or any other form of input data.} and labels $(x_b, y_b) \sim \mathcal{D}_{tr}^b$ from some predetermined set of base classes, where $\mathcal{D}_{tr}^b$ is a training distribution and $x_b$ and $y_b$ are, respectively, images and labels from the $N_b$ base classes $C_b = \{1 \ldots N_b\}$ of our training set. After performing some form of pretraining using $\mathcal{D}_{tr}^b$, we are given a second set of training images and labels $(x_n, y_n) \sim \mathcal{D}_{fs}^n$, with all labels from a set of $N_n$ novel classes $C_n = \{N_b + 1, \ldots, N_b + N_n\}$ which are disjoint from the previously-learned base classes (\ie $C_b \cap C_n = \emptyset$). Importantly, the number of samples $(x_n, y_n)$ provided in this stage is extremely small, often as low as five or even just one. Finally, performance is measured by the classification accuracy of the final system on a test set of samples from the novel classes, $(x_n, y_n) \sim \mathcal{D}_{ts}^n$. In the following text, we first discuss some prerequisites in Sections \ref{sec:cos} and \ref{sec:dae}, then provide a detailed description of our method in Sections \ref{sec:sdnn} and \ref{sec:noise}.

\subsection{Cosine Classifiers \label{sec:cos}}

Cosine Classifiers are one widely used approach for few shot learning, originally explored by \cite{cosine1} and \cite{cosine2}. In this context, few shot learning is performed by initially training a neural network feature extractor $f = F(x)$ on the data $(x_b, y_b) \sim \mathcal{D}_{tr}^b$ as described above, as well as a set of weight vectors $W_b = [w_1, \ldots, w_{N_b}]$. The critical difference between Cosine Classifier pretraining and standard supervised training comes in the per-class logit computation. In standard supervised training, the logits for class $i$ would be computed as the dot product between the weights and the embedding: $z_i = f^T w_i$. In contrast, a Cosine Classifier computes the logits as the cosine distance between the weights and the features:

\begin{equation}
z_i = \cos(f, w_i) = \frac{f^T w_i}{\lVert f \rVert \lVert w_i \rVert}
\end{equation}

\noindent which has the advantage of producing features with reduced intra-class variance. 

The logits are then converted into probabilities $p_i$ using the softmax function $p_i = \exp(\gamma z_i) / \sum_{j \in C_b} \exp(\gamma z_j)$ where $\gamma$ is a learned inverse temperature parameter. Given these final predictions $p_i$ for each class, the network is trained by optimizing standard cross entropy loss:

\begin{equation}
\mathcal{L}(p) = \mathbb{E}_{(x, y) \sim \mathcal{D}_{tr}^b} \left[-\log p_{y} \right].
\label{eq:ce}
\end{equation}

During the novel-class training stage, the weight matrix $W_n$ for novel classes is computed as $w_i = \bar{z}_i, \forall i \in C_n$, where $\bar{z}_i$ is the average over all values of $z_i$ for all images $x_n$ in the novel class training set with label $y_n = i$. In the final inference stage, the new matrix $W_n$ is used in place of $W_b$ to compute probabilities as described above. In other words, the extracted features of the novel classes are directly averaged across samples of the same class and used as weights for further classification.

\subsection{Denoising Autoencoders \label{sec:dae}}

Originally introduced and explored by Vincent \etal \cite{Vincent2008ExtractingAC}, Denoising Autoencoders (DAEs) are a form of autoencoding neural network designed to improve feature robustness by reconstructing a given feature vector into a more likely configuration. A DAE $r(\cdot)$ operates on a feature vector $\hat{f} = g(f)$ that his been corrupted from its original form $f$ with some type of noise $g(\cdot)$ (\eg additive Gaussian noise) by attempting to construct a new feature vector $r(\hat{f})$ which is as close as possible to the original feature vector $f$.

Alain \etal \cite{Alain2014WhatRA} show that, in the case of a DAE trained with additive Gaussian noise, preforming inference on noiseless input will actually cause the DAE to estimate the gradient of the density function of its input, and as such the vector $(r(f) - f)$ will point towards more likely configurations of $f$, \ie the manifold of the input data. Previously, Gidaris \etal \cite{Gidaris_DAE} made use of this fact to refine the novel feature weights $W_n$ using DAEs and improve few shot performance. We use this work as inspiration for our own. 

\subsection{Self-Denoising Neural Networks \label{sec:sdnn}}

\begin{figure*}
\centering
\includegraphics[width=0.75\linewidth]{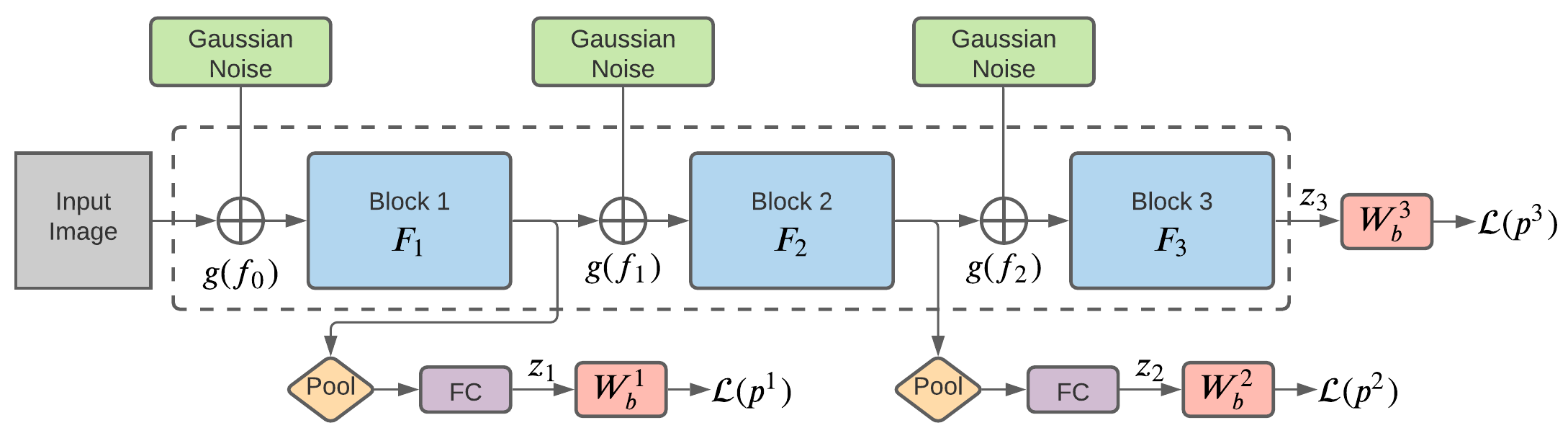}  
\caption{A sample SDNN architecture with Gaussian noise. During training, the features after each block are perturbed using additive Gaussian noise. After each block, the features are pooled and passed through an FC layer so that auxiliary losses $\mathcal{L}(p^l)$ can classify the modified features. During inference, no noise is added, and the weights $W^l_b$ are replaced with the weights  $W^l_n$.}
\label{fig:sdnn}
\end{figure*}

We take the above observations about DAEs and use them to expand existing Cosine Classifier-based few shot learners in a different direction from previous research. Rather than add an additional DAE module to the end of the network as was done in \cite{Gidaris_DAE}, we treat the entire existing network as a set of denoising modules, making minimal architectural changes and achieving considerable performance improvements over simply using a DAE. 

To define this new architecture, we first break our feature extraction network $F$ into multiple blocks $F_1, F_2, F_3$ \footnote{For this example, we use only 3 blocks because that corresponds to the WideResNet \cite{wideRN} network featured most prominently in our experiments section. Other networks, such as ResNet10 \cite{resnet} or Inception \cite{inception} might be broken up into differing numbers of blocks.}. Many modern architectures such as ResNet \cite{resnet} are already naturally organized into blocks, making this division extremely straight-forward. Rather than simply injecting noise at the end and having an additional DAE module denoise the features $f = F(x) = F_3(F_2(F_1(x)))$ produced by the full network, we instead treat each block $F_l$ as its own denoising module by injecting noise at the beginning of each block and denoising the signal before it reaches the next one. 

Specifically, given features $f_{l-1} \in \mathbb{R}^{h_{l-1} \times w_{l-1} \times c_{l-1}}$ (where $h_l$, $w_l$ and $c_l$ represent the height, width, and channels of the feature map at layer $l$) produced from block $F_{l-1}$, we compute $f_l$ during initial training as $f_l = F_l(g(f_{l-1})) \in \mathbb{R}^{h_{l} \times w_{l} \times c_{l}}$, where $g(\cdot)$ is a generic noise function that perturbs its input with any of several more specific noise functions, as discussed below in Section \ref{sec:noise}. This modification is motivated in part by the observation made in Section \ref{sec:dae} that a DAE naturally adds a vector to the input features which points it towards the data manifold. Our hope is to produce a layer that projects towards the data manifold and processes the features at the same time.

Ideally, we would like $f_l$ to produce an input that is robust to the noise that was added to $f_{l-1}$. Applying a standard DAE structure would mean reconstructing $f_{l-1}$ from $g(f_{l-1})$, but this would require either adding additional layers to create a DAE, or enforcing that $f_l$ matches $f_{l-1}$. Since our goal is to retain the original architecture, both of these solutions are undesirable.

Instead, rather than force the network to explicitly denoise $f_{l-1}$, we enforce that $f_l$ remains useful for classification despite the noise. To do this, we add an additional auxiliary classification loss after each block. More specifically, given an uncorrupted feature $f_l \in \mathbb{R}^{h_{l} \times w_{l} \times d_{l}}$, we perform

\begin{align}
\begin{split}
z^l_i =& \cos (\text{FC}(\text{MaxPool}(f_l)), w^l_i) \\
p^l_i=& \text{softmax}(\gamma z^l_i)
\label{eq:pool}
\end{split}
\end{align}

\noindent where FC is a fully-connected layer, and MaxPool is a max-pooling operation across the spatial dimensions of $f_l$. The final predictions $p^l_i$ are then learned using $\mathcal{L}(p^l_i)$, the same cross entropy loss defined in equation \ref{eq:ce}. The weights $w^l_i$  are similar to the weights $W_b$ described in Section \ref{sec:cos}, however they are expanded to include a layer index $W^l_b = [w^l_1, \ldots, w^l_{N_b}]$; in other words, the auxiliary classifiers function identically to the classifier at the end of the unmodified network. This constrains each block to produce the same logits as one another, enforcing the requirement that the network produces features meaningful for classification at every level of the network. Auxiliary losses of this style have been used in classification problems for a long time, \eg when training Inception \cite{inception}. The application of noise in previous layers, however, expands the purpose of these additional losses beyond gradient flow and regularization, using them to counteract the added input noise. As we show empirically in Section \ref{subsec:ablation}, these losses are necessary to see gains from denoising. See Figure \ref{fig:sdnn} for an illustration of the full architecture.

During the novel class learning stage, noise is no longer added and the new weights $W^l_n$ are constructed in the same way as the weights for the Cosine Classifier by assigning $w^l_i = \bar{z}^l_i, \forall i \in C_n$, where $\bar{z}^l_i$ is the average over all values of $z^l_i$ for all images $x_n$ in the novel class training set with label $y_n = i$. The final predictions $p_i$ are computed from the average of all predictions at each layer: $p_i = \sum_l p^l_i$, where $l$ indexes over the number of blocks in the network (either 2 or 3 in all experiments within this paper).

In summary, we modify existing neural network training schemes by repeatedly adding noise to the network while forcing it to produce the same features despite the added noise. Thus, though the network is not performing a true denoising reconstruction, it is still developing a meaningful understanding of the feature space, and when the noise is removed during the novel training and inference stages the network is still able to implicitly refine the previous stage's features into a more likely configuration. 

As a final note we discuss the similarities of our method with stacked DAEs \cite{Vincent2010StackedDA}. Both approaches involve repeated denoising, however we emphasize certain important differences. First of all, stacked DAEs are trained in stages, requiring specific architectures that may require need ot be carefully crafted. Our method, on the other hand, can be easily applied to any existing network architecture just by adding individual pooling and fully connected layers in a few key locations, making it much easier to train and introducing extremely few new hyperparameters. Our method also trains in a single pass using the same hyperparameters as its base architecture, versus the multi-stage training of stacked DAEs. Finally, at a more technical level, our method does not actually autoencode anything; the features are only denoised in the sense that they attempt to produce the same classification outputs as noiseless features. There is no actual reconstruction that occurs.

\subsection{Types of Noise \label{sec:noise}}

The noise function $g(\cdot)$ described in Section \ref{sec:sdnn} can take any number of forms. In this paper, we will explore two different constructions: Dropout noise and Additive Gaussian noise. 

\textbf{Dropout Noise:} Dropout noise is modelled after the widely used method of Dropout \cite{dropout}. Dropout works by first generating a random binary mask $m\in \mathbb{R}^{h \times w \times c}$, where $m_{i,j,k} = 0$ with probability $p_{\text{drop}}$, and then computing $g(f) = f \odot m$, where $\odot$ is an element-wise product. We modify dropout by making it non-spatial in the same manner described below for Gaussian noise, \ie sampling $m\in \mathbb{R}^c$ and broadcasting it into the spatial dimensions.

\begin{figure}
  \centering
\begin{subfigure}{.23\textwidth}
  \centering
  \includegraphics[width=0.8\linewidth]{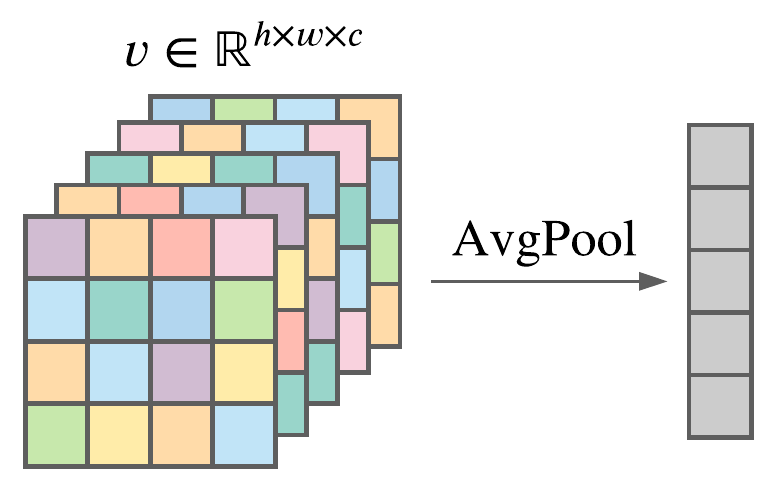}  
  \caption{Spatial Noise}
  \label{fig:spat_noise}
\end{subfigure}
\begin{subfigure}{.23\textwidth}
  \centering
  \includegraphics[width=1.0\linewidth]{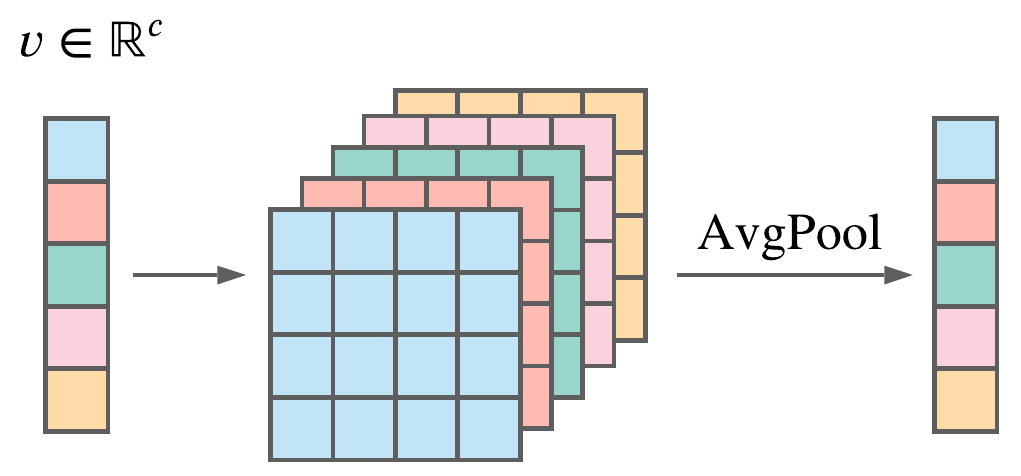}  
  \caption{Non-Spatial Noise}
  \label{fig:nonspat_noise}
\end{subfigure}
\caption{Comparison between spatial and non-spatial noise. (a) Spatial noise perturbs each location in the feature map differently, and therefore averages out to 0 when pooled. (b) Non-spatial noise is constant over each location, and can therefore be pooled without changing its values.}
\label{fig:noise}
\end{figure}

\noindent \textbf{Additive Gaussian Noise:} An additive Gaussian noise function $g(\cdot)$ is defined simply as $g(f) = f + v$ where $v \in \mathbb{R}^c$ and  $v_i \sim \mathcal{N} (0, \sigma)$ for some standard deviation parameter $\sigma$, with $i \in \{1, \ldots c\}$ indexing the $c$ channels of $f$ .   Note that we have  $f \in \mathbb{R}^{h \times w \times c}$, and we are thus implicitly embedding the vector $v \in \mathbb{R}^c$ into  $\mathbb{R}^{h \times w \times c}$ by copying values across the spatial dimensions so that the addition may be performed properly. This operation is known in popular tensor-programming libraries like Pytorch \cite{pytorch} as ``broadcasting''. This formulation, which we call ``non-spatial'' noise sampling, is an important subtlety in the construction of SDNNs. See Figure \ref{fig:noise} for an illustration of this procedure. 

To see why we must use non-spatial noise, first make the simplifying assumption that the pooling layer in equation \ref{eq:pool} were an average pooling layer instead of a max pooling layer, and that the network block $F$  was a simple linear layer with weights $\phi$. In this case, if $v$ were not 1-dimensional and each index of $v \in \mathbb{R}^{h \times w \times c}$ were sampled separately, then for $F(g(f))$ we would have $F(g(f)) = \phi(f + v)$, and therefore

\begin{align}
\begin{split}
\text{AvgPool}(F(g(f)) & =\phi \frac{1}{hw}\sum_{(i,j)}^{h,w} (v_{i, j} + f_{i, j}) \\
& =\phi \frac{1}{hw}\sum_{(i,j)}^{h,w} v_{i, j} + \phi \frac{1}{hw}\sum_{(i,j)}^{h,w} f_{i, j} \\
& \approx \phi \frac{1}{hw}\sum_{(i,j)}^{h,w} f_{i, j}
\end{split}
\end{align}

\noindent where the last line occurs because $\mathbb{E} (v_{i,j}) = 0$. Thus, in order for the noise to have any effect, it must not be sampled over the spatial component. As we see in Section \ref{subsec:ablation}, non-linearities in the network blocks and max pooling layers mitigate some of these issues, but are not sufficient to completely eliminate the disadvantages of using spatial noise. See Figure \ref{fig:noise} for an illustration of this point.

\section{Experiments \label{sec:exp}}

In this section, we will first describe the implementation details of SDNN training in Section \ref{subsec:impl}. We then describe the datasets and metrics used for evaluation in Section \ref{subsec:data} and compare our method against contemporary methods in Section \ref{subsec:sota_comp}. In Section \ref{subsec:ablation} we perform ablation experiments to provide a detailed analysis of our method. Finally, in Section \ref{subsec:action} we evaluate our method on the task of few shot learning for human action detection in video.

\subsection{Implementation Details}
\label{subsec:impl}

One of the most significant advantages of SDNNs is that they can be trained making only minimal modifications to existing networks. As such, our training hyperparameters for image classification mimic those of \cite{Gidaris_2019_ICCV}. More specifically, we perform our 2D image experiments using the WideResNet-28-10 (WRN-28-10) architecture \cite{wideRN}. We perform optimization using Stochastic Gradient Descent (SGD) for 26 epochs, with an initial learning rate of $0.1$ and reducing by a factor of 10 after epoch 20 and again after epoch 23. The inverse softmax temperature hyperparameter $\gamma$ is initialized to 10. 

When training the WRN-28-10 backbone, we perform self-denoising on all three residual blocks, adding noise before each block and an auxiliary block afterwards. When pooling, we pool to a spatial size of $2 \times 2$ and concatenate the features at all four spatial locations before feeding them into the FC layer. Except where otherwise noted, all SDNN experiments are performed using either additive Gaussian noise with $\sigma=0.06$ or Dropout with probability $0.1$. During inference, no noise is added anywhere in the network, and the logits computed from each auxiliary loss are averaged to make the final predictions. 

\subsection{Image Data and Evaluation Metrics}
\label{subsec:data}

We perform all our main experiments on three standard datasets used for few shot learning: \textit{mini}ImageNet \cite{miniimagenet}, \textit{tiered}ImageNet \cite{tin}, and CIFAR-FS \cite{cifar}.

\noindent\textbf{\textit{mini}ImageNet} consists of 100 classes randomly picked from the ImageNet dataset \cite{ILSVRC15} with 600 images of size 84$\times$84 pixels per class. We follow the exact same setup as \cite{Gidaris_2019_ICCV} and others, using 64, 16 and 20 classes as our training, validation and test classes respectively. Also following \cite{Gidaris_2019_ICCV} for consistency, we resample each image to $80 \times 80$ before feeding it into the network.

\noindent\textbf{\textit{tiered}ImageNet} consists of 608 classes randomly picked from ImageNet \cite{ILSVRC15}. It consists of 779,165 images in total, all of resolution 84$\times$ 84 pixels. We use 351, 97 and 160 classes in our training, validation and test classes respectively, and, again similar to \cite{Gidaris_2019_ICCV}, we resample each image during training to $80 \times 80$.

\noindent\textbf{CIFAR- FS} is a few-shot dataset created by dividing the 100 classes of CIFAR-100 into 64 base classes, 16 validation classes, and 20 novel test classes. There are 60000 images in total in this dataset, each with a resolution of 32$\times$ 32 pixels.

\noindent\textbf{Evaluation Metrics: } Following standard few shot classification algorithm practices, our experiments are evaluated based on classification accuracy averaged over a large number of episodes. To be more precise, each episode is a $N_n$-way $K$-shot problem where $K$ samples are selected at random from $N_n$ randomly-selected novel classes of the parent dataset. The $K$ samples form a support set which the network uses to guide inference. $M$ images are then chosen from the $N_n$ chosen classes to form a test set. The classification accuracy is computed over the $M \times N_n$ images. The final few shot scores are computed as the 95\% confidence interval of the accuracy over all the episodes. Except where otherwise noted, we use $N_n = 5$, $M = 15$, and $K = 1$ or $K=5$ (labelled as ``$K$-Shot'' in the appropriate tables).

 \begin{table*}
\centering
\footnotesize
\resizebox{\linewidth}{!}{
\begin{tabular}{@{}lc|ccccccc}
                      & Backbone  & \multicolumn{2}{c}{\textit{mini}ImageNet}                         & \multicolumn{2}{c}{CIFAR-FS}                              & \multicolumn{2}{c}{\textit{tiered}ImageNet}                               &           \\
                      &           & \multicolumn{1}{c}{1-Shot}  & \multicolumn{1}{c}{5-Shot}  & \multicolumn{1}{c}{1-Shot}  & \multicolumn{1}{c}{5-Shot}  & \multicolumn{1}{c}{1-Shot}  & \multicolumn{1}{c}{5-Shot}          &           \\ \midrule%
Baseline (Cosine) \cite{cosine1, cosine2}    & WRN-28-10 & 58.46   $\pm$ 0.45\%         &         75.45   $\pm$ 0.34\%               & 72.63 $\pm$ 0.49\%                     & 85.69 $\pm$ 0.34\%                      & 67.46 $\pm$ 0.51\%          & 82.786   $\pm$ 0.37\%       \\

Shot-Free\cite{Ravichandran2019FewShotLW} & ResNet-12& 59.04 $\pm$ n/a&77.64 $\pm$ n/a&69.2 $\pm$ n/a&84.7 $\pm$ n/a&63.5 $\pm$ n/a &82.59 $\pm$ n/a \\
TEWAM\cite{Qiao2019TransductiveEA} & ResNet-12& 60.07 $\pm$ n/a&75.90 $\pm$ n/a&70.4 $\pm$ n/a&81.3 $\pm$ n/a &- &-\\
MetaOptNet\cite{Lee2019MetaLearningWD} & ResNet-12& 62.64 $\pm$ 0.61\%& 78.63 $\pm$ 0.46\%&  72.6 $\pm$ 0.70\%& 84.3 $\pm$ 0.50\% &65.99 $\pm$ 0.72\%&81.56 $\pm$ 0.53\%\\
Fine-tuning\cite{Dhillon2020ABF}&WRN-28-10& 57.73 $\pm$ 0.62\%  & 78.17 $\pm$ 0.49\%& - & - & 66.58 $\pm$ 0.70\% & 85.55 $\pm$ 0.48\%\\
Qiao \etal \cite{Qiao2018FewShotIR}    & WRN-28-10 & 59.6 $\pm$ 0.41\%           & 73.74 $\pm$ 0.19\%          & -                           & -                           & -                           & -                                   &           \\
LEO \cite{Rusu2019MetaLearningWL}           & WRN-28-10 & 61.76 $\pm$ 0.08\%          & 77.59 $\pm$ 0.12\%          & -                           & -                           & 66.33 $\pm$ 0.05\%          & 81.44 $\pm$ 0.09\%       \\
DAE \cite{Gidaris_DAE}           & WRN-28-10 & 61.07 $\pm$ 0.15\%          & 76.75 $\pm$ 0.11\%          & -                           & -                           & 68.18 $\pm$ 0.16\%          & 83.09 $\pm$ 0.12\%        \\
BF3S (CC+Rot) \cite{Gidaris_2019_ICCV} & WRN-28-10 & 62.93 $\pm$ 0.45\%          & 79.87 $\pm$ 0.33\%          & 73.62 $\pm$ 0.31\%          & 86.05 $\pm$ 0.22\%          & 70.53 $\pm$ 0.51\%          & 84.98 $\pm$ 0.36\%        \\
\midrule
Dropout SDNN (Ours)           & WRN-28-10 & 61.52 $\pm$ 0.45\%          &      78.25 $\pm$ 0.33\%     &  74.98 $\pm$ 0.30\%       &    87.00 $\pm$ 0.21\%  &   69.50 $\pm$ 0.50\%      &    84.34 $\pm$ 0.35\%     \\
Gaussian SDNN (Ours)           & WRN-28-10 & 62.12 $\pm$ 0.45\%          & 78.97 $\pm$ 0.32\%          & 75.31 $\pm$ 0.30\%          & \textbf{87.40 $\pm$ 0.22\%} & 69.29 $\pm$ 0.50\%          & 84.53 $\pm$ 0.35\%          \\
Gaussian SDNN + BF3S (Ours)     & WRN-28-10 & \textbf{64.74 $\pm$ 0.45\%} & \textbf{81.47 $\pm$ 0.30\%} & \textbf{75.60 $\pm$ 0.30\%} & 87.30 $\pm$ 0.22\%          & \textbf{71.40 $\pm$ 0.50\%} & \textbf{85.90 $\pm$ 0.34\%}\\
\bottomrule
\end{tabular}
}
 \caption{Confidence intervals for 5-way classification of several methods, including our own. Inference is performed by sampling images from 5 different novel classes for 2000 iterations on the \textit{mini}ImageNet and \textit{tiered}ImageNet datasets, and 5000 iterations on the CIFAR-FS dataset. ``Dropout SDNN''  and ``Gaussian SDNN'' represent our vanilla method as described in Section \ref{sec:noise} using either Dropout noise or additive Gaussian noise, whereas ``Gaussian SDNN + BF3S'' is a naive combination of our method with the BF3S method \cite{Gidaris_2019_ICCV} intended to show how our method can be effectively combined with others.}

\label{tab:main_res1}
    \end{table*}

\subsection{Image Evaluation Results}
\label{subsec:sota_comp}
In this section, we compare our proposed SDNN approach against contemporary methods on \textit{mini}ImageNet, CIFAR-FS and \textit{tiered}ImageNet.  Table \ref{tab:main_res1} lists our performance for each of these datasets, performing inference for $2000$, $5000$ and $2000$ iterations on \textit{mini}ImageNet, CIFAR-FS and \textit{tiered}ImageNet respectively in an episodic fashion by randomly sampling $5$ novel classes per episode. ``Baseline (Cosine)'' is a re-implementation of \cite{cosine1} using a WideResNet-28-10 \cite{wideRN} backbone. This re-implementation is taken directly from the publicly-available code-base for \cite{Gidaris_2019_ICCV} and has not been modified.

We observe that our method by itself with either Dropout or additive Gaussian noise out-performs the baseline on all three datasets by a significant margin, with additive Gaussian noise performing better than Dropout. An unmodified version of our method also perform favorably against many prior methods, even achieving state-of-the-art performance on the ``CIFAR-FS'' dataset without modification. Notably, our method out-performs the DAE \cite{Gidaris_DAE} method on all four experiments in which they are comparable. The DAE method itself was trained using an extremely similar set of hyperparameters and architectures as our baseline. This indicates the superiority of our method despite its simplicity - the DAE method required the training of a custom graph-neutral network architecture in a separate stage of pretraining. Our method, by comparison, did not require any additional training stages.

\noindent\textbf{SDNN + BF3S}: Table \ref{tab:main_res1} also contains an additional experiment, labelled ``Gaussian SDNN + BF3S'', which combines a Gaussian SDNN with the rotation prediction task described in \cite{Gidaris_2019_ICCV}. Specifically, BF3S modifies the pretraining procedure by rotating each input image three times in 90-degree increments before feeding all four images into the network. The network is then augmented with an additional residual block whose output is used to train a four-way classifier that determines which of the four possible rotations ($0^{\circ}, 90^{\circ}, 180^{\circ}$, or $270^{\circ}$) were preformed on the input. Success in this auxiliary task requires the network to develop richer features beyond those needed for classification, and as such improves the quality of extracted features in downstream tasks.

The BF3S architecture is modified into an SDNN in the same way as the vanilla WRN-28-10 architecture - auxiliary losses are added before each block, and Gaussian noise with $\sigma=0.06$ is added afterwards. 

Because we believe this rotation classification task is essentially orthogonal to the SDNN methodology, we are motivated to include this experiment in order to demonstrate that existing few shot methods can be easily and effectively turned into SDNNs. Indeed, we observe from Table \ref{tab:main_res1} that ``SDNN + BF3S' out-performs all other methods, most importantly the ``BF3S'' method, supporting our assertion that SDNNs are a broadly applicable technique that will not interfere when added with other, orthogonal techniques.

\subsection{Ablation Experiments}

\label{subsec:ablation}
In this section we analyze the various components of our proposed approach. We report all of our experimental results on the \textit{mini}ImageNet dataset unless otherwise specified. For all experiments in this section, inference is performed for 2000 episodes by sampling images from 5 different novel classes unless otherwise specified.

\begin{table}[ht]
\centering
\footnotesize
\begin{tabular}{lc|ll}
\multicolumn{1}{c}{\textbf{Aux?}} & \multicolumn{1}{c|}{\textbf{Noise}} & \multicolumn{1}{c}{\textbf{1-Shot}} & \multicolumn{1}{c}{\textbf{5-Shot}} \\ \hline
No                                & -                                    & 58.43 $\pm$ 0.45\%                  & 75.45 $\pm$ 0.34\%                  \\
Yes                               & -                                    & 60.91 $\pm$ 0.45\%                  & 77.84 $\pm$ 0.33\%                  \\
No                                & Dropout                              & 58.15 $\pm$ 0.44\%                    & 76.10 $\pm$ 0.33\%                    \\
No                                & Gaussian                             & 58.73 $\pm$ 0.45\%                    & 76.16 $\pm$ 0.34\%       \\             
Yes                                & Gaussian                             &  \textbf{62.12 $\pm$ 0.45\%}         &   \textbf{78.69 $\pm$ 0.32\%}    
\end{tabular}
\caption{Confidence intervals for 5-way classification accuracy applying additive Gaussian noise and Dropout noise with and without auxiliary losses on the \textit{mini}ImageNet dataset with a WideResNet backbone. Inference is performed by sampling images from 5 different novel classes for 2000 iterations. ``Aux?'' indicates if auxiliary losses were used, and ``Noise'' indicates what type noise was used.}
\label{tab:aux_noise}
\end{table}

\noindent \textbf{Auxiliary Losses and Noise:} SDNNs make essentially two modifications to a standard neural network training procedure: the addition of noise and the training of auxiliary losses. In Table \ref{tab:aux_noise}, we perform the critical experiment of removing each of these components in turn. In doing so, we make two observations. The first observation is that while auxiliary losses are helpful to performance on their own, the addition of noise is critical for achieving the best results, as seen from, for instance, the 1-shot performance increasing from $60.91$ to $62.12$ as Gaussian noise is added. 

Our second observation is even more significant: additive Gaussian and Dropout noise have a very small effect on performance if they are not paired with some sort of auxiliary loss - achieving $58.73$ 1-Shot performance as opposed to $58.43$. This has significant ramifications - it confirms that the performance gains from using an SDNN exist only because of the specific combination of noise with auxiliary losses. In fact, without the auxiliary losses, the network learns to mostly ignore the noise. We hypothesize that this occurs because adding noise only perturbs the features a very small amount, and if the network is allowed more layers to process the noisy features before classification, the effect of the noise will be mitigated. 

\begin{table}[ht]
\centering
\footnotesize
\begin{tabular}{lc|ll}
Noise Type & Spatial? & \multicolumn{1}{c}{\textbf{1-Shot}} & \multicolumn{1}{c}{\textbf{5-Shot}} \\ \hline
None & -                       & 60.91 $\pm$ 0.45\%                  &  77.84 $\pm$ 0.33\%               \\
Dropout   & Yes                  & 61.34 $\pm$ 0.45\%       &   78.18 $\pm$ 0.32\%        \\
Gaussian   & Yes                    & 61.38 $\pm$ 0.45\%                  & 78.23 $\pm$ 0.32\%                  \\
Gaussian  (AVG) & Yes                    & 60.68 $\pm$ 0.45\%                  &        77.69 $\pm$ 0.33\%           \\
Dropout    & No               & 61.52 $\pm$ 0.45\%         &     78.25 $\pm$ 0.33\%     \\
Gaussian         & No          & \textbf{62.12 $\pm$ 0.45\%}         &   \textbf{78.69 $\pm$ 0.32\%}        \\
Gaussian  (AVG) & No                    & 61.47 $\pm$ 0.44\%                &        78.17 $\pm$ 0.32\%  
\end{tabular}
\caption{Confidence intervals for 5-way classification accuracy sampling either spatial ($v \in \mathbb{R}^{w \times h \times c}$) or non-spatial ($v \in \mathbb{R}^{c}$) noise, as described in Section \ref{sec:noise}. ``AVG'' stands for the use of average pooling instead of max pooling in our implementation of equation \ref{eq:pool}. Inference is performed by sampling images from 5 different novel classes for 2000 iterations on \textit{mini}ImageNet.}
\label{tab:spat_non_spat_noise}
\end{table}

\noindent \textbf{Spatial vs. Non-Spatial: } Next, we justify our statements in Section \ref{sec:noise} regarding the need for noise to be non-spatial (\ie to select a noise vector $v \in \mathbb{R}^c$ as opposed to $v \in \mathbb{R}^{w \times h \times c}$). Table \ref{tab:spat_non_spat_noise} shows what happens when noise is added with spatial and non-spatial schemes. Our first observation is that, as expected, non-spatial noise out-performs spatial noise. In the case of additive Gaussian noise, we see an improvement from $61.38$ to $62.12$ in 1-Shot performance. The improvement for Dropout noise is much smaller to the point of statistical insignificance, improving from $61.43$ to $61.52$, suggesting that the non-spatial constraint may not be as necessary for Dropout. This could be anticipated, since the arguments for non-spatial noise in Section \ref{sec:noise} do not necessarily apply to Dropout noise.

In Section \ref{sec:noise} we motivated the need for non-spatial noise by illustrating that spatial noise, when pooled, will have very nearly zero effect. The reason we still see spatial noise effecting performance is that in an SDNN the features are not actually pooled until after they pass through a residual block, which might not act linearly across all of the noise. Additionally, we note that these experiments were performed using max pooling (equation \ref{eq:pool}) as opposed to average pooling, which reacts differently as an operation to $0$-mean additive Gaussian noise. We therefore include an additional experiment in Table \ref{tab:spat_non_spat_noise}, labelled ``Gaussian (AVG)'' which uses average pooling. As expected, we find no statistically significant difference between spatial Gaussian noise with average pooling and no noise at all. For completeness, we also include results for non-spatial Gaussian noise with average pooling, which confirm that the loss in performance with spatial average pooled noise is not entirely due to the choice of pooling method.

\begin{table}[ht]
\centering
\footnotesize
\begin{tabular}{lll|ll}
\textbf{G1} & \textbf{G2} & \textbf{G3} & \multicolumn{1}{c}{\textbf{1-Shot}} & \multicolumn{1}{c}{\textbf{5-Shot}} \\ \hline
            &             &             & 60.91 $\pm$ 0.45\%                  & 77.84 $\pm$ 0.33\%                  \\
\checkmark  &             &             & 61.25   $\pm$ 0.45\%                & 78.13 $\pm$ 0.32\%                  \\
            & \checkmark  &             & 61.26   $\pm$ 0.45\%                & 78.05 $\pm$   0.33\%                \\
            &             & \checkmark  & 61.74   $\pm$ 0.45\%                & 78.46 $\pm$   0.32\%                \\
\checkmark  & \checkmark  &             & 61.26   $\pm$ 0.45\%                & 78.07 $\pm$ 0.32\%                  \\
\checkmark  &             & \checkmark  & 61.38   $\pm$ 0.44\%                & 78.34 $\pm$ 0.32\%                  \\
            & \checkmark  & \checkmark  & 61.65   $\pm$ 0.44\%                & 78.57 $\pm$ 0.32\%                  \\
\checkmark  & \checkmark  & \checkmark  &  \textbf{62.12 $\pm$ 0.45\%}         & \textbf{78.69 $\pm$ 0.32}\%                           
\end{tabular}
\caption{Confidence intervals for 5-way classification accuracy applying Gaussian noise at different locations on the \textit{mini}ImageNet dataset with a WideResNet backbone. Inference is performed by sampling images from 5 different novel classes for 2000 iterations. The columns marked ``G1'' through ``G3'' indicate whether or not Gaussian noise was added to blocks 1 through 3, respectively.}
\label{tab:noise_loc}
\end{table}

\noindent \textbf{Which Layers Need Noise?} In this section we try to analyze at which point in the network the denoising should be performed. We use the Cosine Classifier with auxiliary losses as our baseline and show all our results in Table \ref{tab:noise_loc}. Here, ``$G_x$'' $(x\in \{1,2,3\})$ indicate adding Gaussian noise to the input of block $x$ of the backbone. Initially, we add auxiliary loss at each block without any feature noise. We then add noise to different blocks in turn. We find that performance increases consistently when adding noise to each block, at least when using WideResNet, but we also consistently observe the biggest performance increase when noise is added to the last block. We believe this occurs because the deeper layers of the network possess stronger semantic information, and are therefore most improved by the addition of noise. In our action detection experiments on deeper networks, this motivates us to perform self-denoising only at the later layers of the network.

\begin{table}[ht]
\centering
\footnotesize
\begin{tabular}{l|ll}
\multicolumn{1}{c|}{\textbf{$\sigma$}} & \multicolumn{1}{c}{\textbf{1-Shot}} & \multicolumn{1}{c}{\textbf{5-Shot}} \\ \hline
0.15                                   & 61.27 $\pm$ 0.44\%                  &  78.14 $\pm$ 0.32\%                  \\    
0.1                                   &  61.53 $\pm$ 0.45\%                  & 78.50 $\pm$ 0.32\%                  \\    
0.08                                   & 61.93 $\pm$ 0.45\%                  & 78.62 $\pm$ 0.32\%                  \\
0.06                                   & \textbf{62.12 $\pm$ 0.45\%}         & \textbf{78.69 $\pm$ 0.32}\%              \\
0.04                                   & 61.82 $\pm$ 0.45\%                  &  78.60 $\pm$ 0.32\%                  
\end{tabular}
\caption{Confidence intervals for 5-way classification accuracy applying additive Gaussian noise at different values of $\sigma$. Inference is performed by sampling images from 5 different novel classes for 2000 iterations on \textit{mini}ImageNet.}
\label{tab:gaussian_noise}
\end{table}

\noindent \textbf{Gaussian Noise Parameters: } Table \ref{tab:gaussian_noise} shows the results of using different levels of Gaussian noise. We observe that the value of $\sigma$ has a significant effect on performance. The network achieves the best performance of $62.12\%$ and $78.69\%$ for 1-shot and 5-shot respectively with a $\sigma$ value of $0.06$. Above $0.06$, the performance drops gradually, as the features become too corrupted and the network is unable to denoise them to the required level. Below $0.06$, we also start to see performance fall as the noise becomes too small to meaningfully alter the network's features.

\begin{table}[ht]
\centering
\footnotesize
\begin{tabular}{l|ll}
\textbf{Prob} & \multicolumn{1}{c}{\textbf{1-Shot}} & \multicolumn{1}{c}{\textbf{5-Shot}} \\ \hline
0.2                   & 60.87 $\pm$ 0.45\%                  & 77.43 $\pm$ 0.33\%                  \\
0.15                   & 61.43 $\pm$ 0.45\%                  & 77.96 $\pm$ 0.33\%                  \\
0.1                   & \textbf{61.52 $\pm$ 0.45}\%                  & \textbf{78.25 $\pm$ 0.33}\%                  \\
0.05                  & 61.42 $\pm$ 0.45\%                  & 78.11 $\pm$ 0.32\%                  \\
0.02                  & 61.24 $\pm$ 0.45\%                  & 78.14 $\pm$ 0.32\%                 
\end{tabular}
\caption{Confidence intervals for 5-way classification accuracy applying Dropout noise at different probabilities of dropping. Inference is performed by sampling images from 5 different novel classes for 2000 iterations on \textit{mini}ImageNet.}
\label{tab:dropout_noise}
\end{table}

\noindent \textbf{Dropout Noise Parameters:} Next, we experiment with dropout noise as described in Section \ref{sec:noise}.
With dropout noise, the model achieves the best performance of $61.52\%$ and $78.25\%$ for 1-shot and 5-shot respectively with a dropout probability of $0.1$. We observe with dropout noise a similar trend to that of additive Gaussian noise. At lower probabilities, there is very little noise to denoise and for higher probabilities, the network is unable to recover the original features effectively.

\subsection{Few Shot Learning for Action Detection \label{subsec:action}}

\begin{table}[]
\centering
\footnotesize
\begin{tabular}{l|c}
                  & \textbf{$nAUDC$ ($\downarrow$)} \\ \hline
Cosine Similarity & 0.745          \\
Gaussian SDNN     & \textbf{0.691}
\end{tabular}
\caption{$nAUDC$ scores on the ActEV SDL Surprise Activity evaluation. A lower score is better.}
\label{tab:actev}
\end{table}

In addition to our experiments in image classification, we also perform an additional evaluation on the task of human action detection in video. Our evaluation is performed on the ``Surprise Activities'' split of the Activities in Extended Video (ActEV) Sequestered Data Leaderboard (SDL) Challenge \cite{actev}, a public competition in which real-time action detection systems are evaluated on security footage in known or unknown facilities. To submit to the ``Surprise Activities'' split, teams must produce a system that can perform training with novel classes online on a remote server. The exact novel classes, as well as the number of samples for training and evaluation, are not made public. SDL submissions are evaluated using the $nAUDC$ metric, computed by plotting the probability of a missed activity detection at temporal-false-alarm rates between $0$ and $0.2$ and calculating the normalized area under the curve. For a more detailed explanation of the ActEV scoring protocol and dataset, see \cite{actev} and the supplementary material.

To perform these experiments on the ActEV SDL Surprise Activities split, we have adapted the action recognition pipeline described in \cite{Gleason2019APS, Gleason2020ActivityDI} to be suitable for the few shot setting. The pipeline itself operates in two stages - a cuboid action proposal stage and an action classification stage - and our modifications are made only to the action classification stage, which is adapted from the I3D architecture \cite{i3d}. The I3D architecture is itself an extension of the InceptionV1 architecture \cite{inception} inflated to use 3D convolutions, and we modify it to be an SDNN by adding Gaussian noise before the ``Mixed4e'' and ``Mixed5b'' layers, adding a pooling and cosine classification layer after the ``Mixed4f'' layer, and replacing the final dot product with a Cosine Classifier as described in Section \ref{sec:cos}. As with our previous experiments in 2D, we make no further changes to the training scheme or hyperparameters of the original method. 

Table \ref{tab:actev} shows the results of an SDNN compared to a simple Cosine Classifier baseline. We see that the SDNN performs better than the baseline, with an $nAUDC$ score of $0.691$ as opposed to $0.745$ (a lower score indicates superior performance). This evaluation shows that our method is effective even for video tasks which use larger datasets and deeper architectures.

\section{Conclusion}

In this paper, we have introduced SDNNs, a novel architecture inspired by the ability of DAEs to make features more robust in a few shot learning setting. We have empirically shown the effectiveness of SDNNs on four different datasets across two different tasks, and performed detailed experimental analysis to motivate our construction.

\section{Acknowledgements}

This research is based upon work supported by the Office of the Director of National Intelligence (ODNI), Intelligence Advanced Research Projects Activity (IARPA), via IARPA R\&D Contract No. D17PC00345. The views and conclusions contained herein are those of the authors and should not be interpreted as necessarily representing the official policies or endorsements, either expressed or implied, of ODNI, IARPA, or the U.S. Government. The U.S. Government is authorized to reproduce and distribute reprints for Governmental purposes notwithstanding any copyright annotation thereon.

{\small
\bibliographystyle{ieee_fullname}
\bibliography{egbib}
}

\clearpage

\section{Supplementary Material}

\subsection{Additional ActEV Information}

In addition to the 328 hours of ground-camera data that was annotated for the ActEV SDL, our training makes use of the extensive additional annotations performed by the public in the MEVA annotation repository \cite{meva}.

For the activity recognition experiments themselves, the conversion of the system into an SDNN uses average pooling down to $1 \times 1$ features in the spatial dimensions, while retaining a depth of 8 in the temporal dimension. The features are then concatenated along the temporal dimension and fed into the fully-connected layers, where a cosine classifier makes the final predictions during training. As in the 2D case, during inference the image feature embeddings from the novel class exemplars are used as classification weights.

\subsection{Additional Implementation Details}

\begin{table}[]
\centering
\begin{tabular}{l|cc}
           Architecture   & \textbf{1-Shot}                              & \textbf{5-Shot}              \\ \hline
3 Blocks, MAX & 75.09 $\pm$ 0.30\%                           & \textbf{87.32 $\pm$ 0.22\%} \\
2 Blocks, AVG & \textbf{75.60 $\pm$ 0.30\%} & 87.30 $\pm$ 0.22\%         
\end{tabular}
\caption{Comparison of the ``Gaussian SDNN + BF3S'' experimental results on the CIFAR-FS dataset. The first row of results uses the SDNN hyperparamters described in the paper (3 blocks with max pooling), and the second row uses those described in this supplementary material (2 blocks and average pooling).}
\label{tab:exp1}
\end{table}

\begin{table}[h]
\centering
\begin{tabular}{l|cc}
                          Architecture             & \textbf{1-Shot}             & \textbf{5-Shot}             \\ \hline
BF3S (CC+Rot) \cite{Gidaris_2019_ICCV} & 70.53 $\pm$ 0.51\%          & 84.98 $\pm$ 0.36\%          \\
3 Blocks, MAX                          & 69.28 $\pm$ 0.50\%          & 83.67 $\pm$ 0.35\%          \\
2 Blocks, AVG                          & \textbf{71.40 $\pm$ 0.50\%} & \textbf{85.90 $\pm$ 0.34\%}
\end{tabular}
\caption{Comparison of the ``Gaussian SDNN + BF3S'' experimental results on the \textit{tiered}ImageNet dataset. The second row of results uses the SDNN hyperparamters described in the paper (3 blocks with max pooling), and the third row uses those described in this supplementary material (2 blocks and average pooling).}
\label{tab:exp}
\end{table}

For two of the experiments in the original paper (the ``Gaussian SDNN + BF3S'' experiment performed on the CIFAR-FS and \textit{tiered}ImageNet dataset in Table 1 of the original paper) our SDNN implementation only applied noise and auxiliary losses to the last two blocks of the network, as opposed to the three used in every other experiment. Additionally, these two experiments were performed with average pooling instead of max pooling. We empirically found that this resulted in better performance on those datasets when combined with rotation classification; see Tables \ref{tab:exp1} and \ref{tab:exp}. In particular, the \textit{tiered}ImageNet performance actually dips below the performance of the unmodified BF3S \cite{Gidaris_2019_ICCV} model when all three blocks are used. We hypothesize that this is because the earlier layers of the network are more susceptible to noise when performing the rotation classification task. 

All of the networks in this paper were each trained on a single Nvidia Titan XP GPU.

\end{document}